\documentclass{article} % For LaTeX2e
\usepackage{iclr2024_conference,times}

\usepackage{amsmath,amsfonts,bm}

\def\eqref#1{equation~\ref{#1}}
\def\1{\bm{1}}

\DeclareMathAlphabet{\mathsfit}{\encodingdefault}{\sfdefault}{m}{sl}
\SetMathAlphabet{\mathsfit}{bold}{\encodingdefault}{\sfdefault}{bx}{n}

\usepackage{hyperref}
\usepackage{url}
\usepackage{inconsolata}
\usepackage{algorithm}
\usepackage{algorithmic}
\usepackage{array}
\usepackage{multirow}
\usepackage{multicol}
\usepackage{caption}
\usepackage{subfigure}
\usepackage{amsfonts} 
\usepackage{booktabs}
\usepackage{amsmath}
\usepackage{amssymb}
\usepackage{graphicx}
\usepackage{wrapfig}
\usepackage{enumitem}
\usepackage{xspace}

\title{Retrieval is Accurate Generation}

\author{Bowen Cao$^{{\spadesuit},}$\thanks{Work done during an internship at Tencent AI Lab.}~~Deng Cai$^{\heartsuit,}$\thanks{Corresponding author.}~~Leyang Cui$^{\heartsuit}$ Xuxin Cheng$^{\spadesuit}$ Wei Bi$^{\heartsuit}$ Yuexian Zou$^{\spadesuit}$ Shuming Shi$^{\heartsuit}$ \\
$\spadesuit$ School of ECE, Peking University, China \\
$\heartsuit$ Tencent AI Lab \\ 
\texttt{\{cbw2021,chengxx\}@stu.pku.edu.cn}\\
\texttt{\{jcykcai,leyangcui,victoriabi,shumingshi\}@tencent.com} \\
}

\iclrfinalcopy % Uncomment for camera-ready version, but NOT for submission.

\begin{document}
\maketitle
\begin{abstract}
Standard language models generate text by selecting tokens from a fixed, finite, and standalone vocabulary. We introduce a novel method that selects context-aware phrases from a collection of supporting documents. One of the most significant challenges for this paradigm shift is determining the training oracles, because a string of text can be segmented in various ways and each segment can be retrieved from numerous possible documents. To address this, we propose to initialize the training oracles using linguistic heuristics and, more importantly, bootstrap the oracles through iterative self-reinforcement. Extensive experiments show that our model not only outperforms standard language models on a variety of knowledge-intensive tasks but also demonstrates improved generation quality in open-ended text generation. For instance, compared to the standard language model counterpart, our model raises the accuracy from 23.47\% to 36.27\% on OpenbookQA, and improves the MAUVE score from 42.61\% to 81.58\% in open-ended text generation. Remarkably, our model also achieves the best performance and the lowest latency among several retrieval-augmented baselines. In conclusion, we assert that retrieval is more accurate generation and hope that our work will encourage further research on this new paradigm shift.
\end{abstract}

\section{Introduction}
%attributable, efficient and more accurate. It is an important direction to explore to make language model more factual.
\textit{Memorization or generalization, that is the question.}

Standard language models (LMs) break down the text generation process into sequential token predictions \citep{mikolov2010recurrent,2020BrownGPT3,chatgpt}. Each token is a word (or sub-word) selected from a fixed, finite, and standalone vocabulary. To make the generation more attributable and accelerate the inference speed, \citet{lan2023cog} propose a method named CoG that retrieves phrases from similar contexts, where the term ``phrase" refers to any contiguous text segments of variable lengths. It is worth noting that, similar to other retrieval-augmented generation frameworks \citep{Li2022ASO,asai2023retrieval}, CoG still employs a two-stage pipeline, specifically document retrieval followed by grounded phrase extraction. The final performance is constrained by the quality and quantity of the return from the first stage. In this paper, we propose a new paradigm that completely removes the dependence on document retrieval. To our best knowledge, our work is the first that performs text generation through direct phrase retrieval.

One core challenge of adopting this novel approach is the construction of the training oracles. That is a function mapping a string of text to an action sequence for creating training examples. For a given text, there exist numerous different ways to segment it into phrases, with each potential phrase being retrievable from a vast array of documents. To better align the generation process and the supporting documents, we introduce a two-fold approach: first, we leverage linguistics-motivated heuristics to initialize the training oracles. Second, we implement a bootstrapping mechanism through iterative self-reinforcement, gradually refining the oracles with each iteration.

Unlike \citet{lan2023cog} which only evaluates the generation fluency in open-ended text generation, we carry out comprehensive and rigorous evaluation in a wide range of knowledge-intensive tasks, \textit{e.g.}, open-domain question answering. Our proposed model exhibits superior zero-shot performance, outperforming the baseline method. For example, on the OpenbookQA dataset, our model dramatically improves upon base LM, presenting an increase in accuracy from 23.47\% to 36.27\% (Table \ref{from pre-trained result}).  Our model also demonstrates improved quality in open-ended text generation, as evidenced by the improvement of 38.97\% in the MAUVE score (Table \ref{open-ended generation}). Moreover, it shows even better performance when switching to an enlarged (Table \ref{enlarged index}) or domain-specific (Table \ref{medical index}) phrase table, without any further training. 
In addition, our model attains the fastest generation speed among retrieval-augmented baselines (Table \ref{open-ended generation}).
We believe that our study can inspire future research to build more efficient and accurate LMs that harness the power of retrieval-based approaches.

In summary, the contributions of this paper can be summarized as follows:
\begin{itemize}[leftmargin=0.6cm]
\setlength{\itemsep}{1pt}
\item We introduce a new approach for language modeling that focuses on directly selecting context-aware phrases from a set of supporting documents.
\item We propose a novel method for decomposing text generation into sequential next-phrase retrieval by linguistics-driven heuristics and iterative self-reinforced bootstrapping.
\item We validate the effectiveness of our models on various downstream tasks, including open-domain and domain-specific question answering, as well as open-ended text generation, highlighting substantial improvements over standard LMs and several retrieval-augmented baselines. 
\end{itemize}
\section{A Unified View of Generation and Retrieval}
\label{sec:unified}
Standard language models (LMs) factorize the generation probability of a sequence $\mathbf{x}= [x_1, x_2, \ldots, x_n]$ into a series of conditional probabilities $p(\mathbf{x}) = \prod_{i=1}^n p(x_i|\mathbf{x}_{<i})$. Hence, the generation is often performed by repeatedly predicting the next token based on the generated sequence thus far (\textit{i.e.}, prefix). The next-token prediction probabilities are computed as
\begin{equation}
    p(x_i |\mathbf{x}_{<i})=\frac{\exp( E_p(\mathbf{x}_{<i})  \cdot E_c(x_i))}{\sum_{x' \in V} \exp( E_p(\mathbf{x}_{<i}) \cdot E_c(x') )},
    % p_{\{\theta,\phi\}}(x_i |\mathbf{x}_{<i})=\frac{\exp( h_\theta(\mathbf{x}_{<i})  \cdot e_\phi(x_i))}{\sum_{x' \in V} \exp( h_\theta(\mathbf{x}_{<i}) \cdot e_\phi(x') )},
    \label{next-token prob}
\end{equation}

\begin{figure*}[t]
\centering
\includegraphics[width=0.95\textwidth]{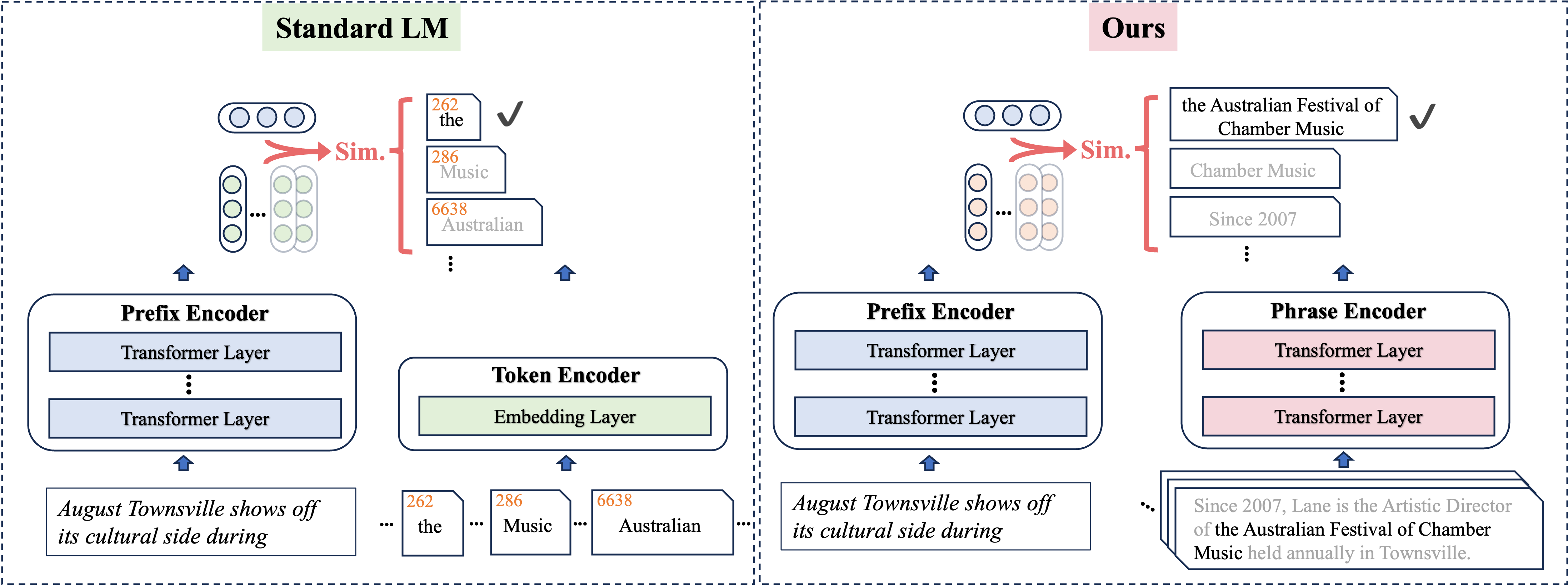} 
\caption{
Comparison between our method and standard language models. Both can be viewed as dual-encoder matching networks connecting source prefixes and target continuations. On the target side, standard language models employ an immediate embedding layer for target tokens from a fixed, finite, and standalone vocabulary. In contrast, our methods uses an expressive phrase encoder for target phrase from an editable, extensible, and contextualized phrase table.}
\label{Fig.overview}
\end{figure*}

\vspace{-5pt}
where $E_p(\mathbf{x}_{<i})$ is a vector representation of the prefix $\mathbf{x}_{<i}$, $E_c(x)$ denotes a vector representation of the token $x$, and $V$ stands for the token vocabulary. 
Through the above notations, we can see that the standard LMs can be viewed as a dual-encoder matching network connecting different prefixes and tokens. Typically, as shown in the left part of Figure \ref{Fig.overview}, the source encoder $E_p$ is implemented by a multi-layer neural network (e.g., Transformers) while the target encoder $E_c$ is simply a token embedding layer. As seen, the design of the dual-encoder network is heavily unbalanced; The source side is much more complex than the target side.

Recently, a retrieval-augmented LM, CoG \citep{lan2023cog}, has been proposed. In addition to token selection, CoG also allows for phrase retrieval (i.e., variable-length $n$-grams) from a collection of supporting documents. From our point of view, CoG augments the target side of conventional LMs. First, the candidate pool is enlarged to include phrases of variable lengths. Second, the target encoder not only considers the candidates themselves but also their contexts.

However, searching phrases from large-scale corpora is resource-intensive. Therefore, CoG adopts a two-stage search strategy: relevant documents are first retrieved to reduce the search space for phrase selection. To construct the training oracles, CoG uses a forward maximum matching algorithm to find the longest matching phrases from the retrieved documents. Despite promising results, CoG cannot guarantee to provide a globally optimal solution for phrase retrieval, and is highly dependent on the external tool for document retrieval. In contrast, we present a new paradigm that generates text directly through phrase retrieval.

\vspace{-4pt}
\section{The Proposed Method}
\vspace{-4pt}
\subsection{Overview}
\vspace{-3pt}
Our research aims to enhance the interpretability and factuality of language models (LMs) by transitioning from token generation to phrase retrieval. First, the semantics of phrases are enhanced by their surrounding contexts \citep{mikolov2013neg1}, leading to a more discriminative representation for inference. Second, each retrieved phrase can be traced back to its original document, enhancing the accountability of the output.
%\footnote{For each output of our model, we can provide a clear lineage or 'account' of where that output came from in terms of the supporting documents.}

To link a given prefix with a set of variable-length phrases, our model follows the dual-encoder structure as described in Section \ref{sec:unified} but emphasizes a balanced design in contrast to standard LMs that heavily favor the source side (see Figure \ref{Fig.overview}). Specifically, the source encoder $E_p(\cdot)$ is a multi-layer neural network (e.g., Transformer) as usual. The target encoder $E_c(\cdot)$ is also a multi-layer neural network to learn context-aware representation for phrases in supporting documents.

Similar to standard LMs, we employ dot product as the matching measure. During inference, we can use efficient maximum inner product search (MIPS) algorithms \citep{shrivastava2014mips1, guo2016mips2,seo2019real} to retrieve from a large pool of candidate phrases. The overall framework is depicted in Figure \ref{Fig.overview}. The remaining question is how to train our models.

\begin{figure*}[t]
\centering
\includegraphics[width=0.95\textwidth]{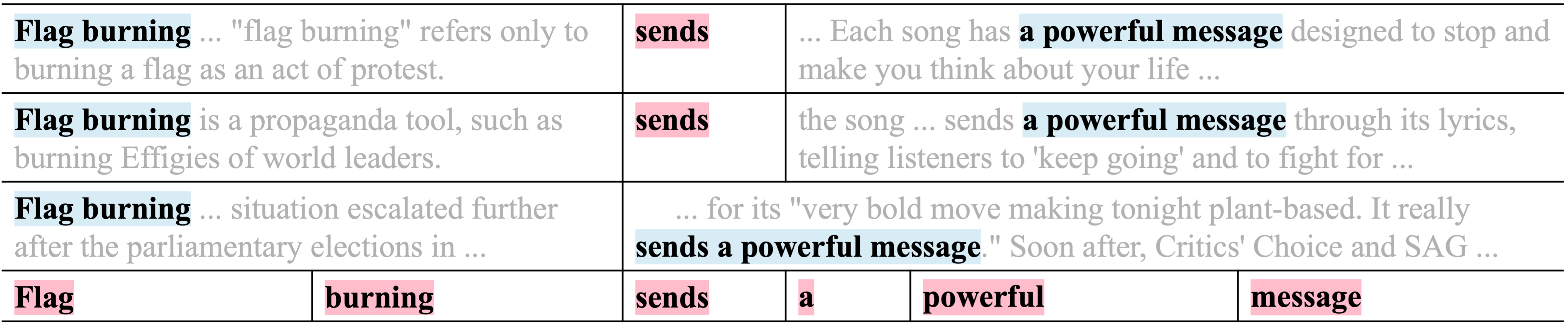} 
\caption{Four possible generation paths for the sentence ``Flag burning sends a powerful message''. Content highlighted in blue (red) are phrases retrieved from supporting documents (from the token vocabulary). Standard LMs can be viewed as only considering the generation path at the bottom.}
\label{Fig.generation_path}
\end{figure*}

% \vspace{-10pt}
\subsection{Training Oracles}
\label{generating training oracles}
% \vspace{-4pt}
We break down text generation into a series of next-phrase retrieval. Formally, each step takes the current prefix $p$ as its state, an oracle policy $\pi^*$ maps the state to an action $\pi^*(p) \rightarrow (f, s)$, where $f$ is a follow-up phrase and $s$ is a copy of the phrase $f$ in a supporting document.

As illustrated in Figure \ref{Fig.generation_path}, to create such triplets $(p, f, s)$ from raw corpora presents two challenges. First, the boundary of the phrase $f$ is unclear given a continuation can be divided in various ways. Second, the source of each phrase $s$ is unclear because a phrase can appear numerous times across a vast number of documents. On the other hand, the variety of generation paths for a given text also indicates that training oracles are crucial for optimal and quick convergence of our models.

To tackle the above problems, we first present a set of linguistics-motivated heuristics to initialize the training oracles (Section \ref{pseudo label}), then describe how we allow the model to refine its generation paths in a self-reinforcement manner (Section \ref{self-reinforcement}).

\vspace{-4pt}
\subsubsection{Linguistics-motivated Heuristics}
\label{pseudo label}
We start to design the training oracles through the following principles.
\vspace{-5pt}
\paragraph{Syntactic Structure.}
Inspired by the syntactic structure of language and its implications on language generation \citep{chomsky1957syntactic,dyer-etal-2016-recurrent,li-etal-2023-explicit}, we restrict the phrase to a contiguous sequence of words that corresponds to a constituent unit in a syntactic parse tree. This approach ensures that each phrase possesses a relatively complete and well-defined meaning, while avoiding arbitrary word combinations that could result in semantic ambiguity or nonsensical formations \citep{morgan1981role}.
\vspace{-5pt}
\paragraph{Distributional Sparsity.}
The inclusion of high-frequency phrases significantly inflates the size of the candidate pool. This is due to our treatment of lexically identical phrases in different contexts as distinct entries in the pool. Consequently, a single high-frequency phrase could potentially introduce tens of thousands, or even millions, of entries. In our analysis of Wikipedia, we discovered that eliminating just the top 1\% of high-frequency phrases could reduce the total number of entries by 50\%. However, these high-frequency phrases, such as 'as well as', often lack specific meanings. Their inclusion may result in imbalanced training, which could adversely affect the model's overall performance. Regarding phrases with extremely low frequency, we consider them to be rare usages with limited practical use. Including them would notably increase the complexity of training. Therefore, we also choose to exclude them.
%Conversely, the top 69\% of lowest frequency phrases only account for 15\% of the total entries.
%Incorporating high-frequency phrases can significantly increase the total number of phrases, leading to an extremely large candidate pool. However, these phrases, such as ``as well as'', often lack specific meanings, and their inclusion may lead to imbalanced training, thereby affecting the overall performance of the model. %Conversely, low-frequency phrases typically represent special entities that display unambiguous meanings across different contexts, rendering contextualized representation redundant.

\vspace{-5pt}
\paragraph{Semantic Similarity.} Although a lexically identical copy of a phrase can be located in various places, it is crucial to account for polysemy \citep{cruse1986polysemy}, as lexically identical phrases can exhibit different meanings depending on their contexts. Moreover, even when lexically identical phrases share similar meanings, subtle nuances can arise from different contexts, necessitating a thorough evaluation of semantic similarity when selecting the most appropriate matching \citep{min2019discrete}.

Specifically, we first run the Stanford Parser\footnote{\url{https://stanfordnlp.github.io/stanza/}} to extract constituents from the training data. We then filter these constituents based on the following criteria: (1) remove trivial constituents with labels such as \texttt{WHADJP}, \texttt{WHADVP}; (2) exclude constituents that are too short ($<2$ words) or too long ($>10$ words); (3) discard constituents with excessively high or low Inverse Document Frequency (IDF) \citep{salton1988idf} values. Notably, we apply a more lenient IDF threshold for longer constituents. Next, we group lexically identical phrases and compute the pairwise semantic similarities using BM25 \citep{robertson2009BM25} and an off-the-shelf phrase encoder \citep{lee2021densephrase}. Consequently, we can identify the most suitable next phrase for each prefix based on the scores. For more detailed information, please refer to the Appendix \ref{Appx: phrase table pruning and phrase matching}.

\vspace{-2pt}
\subsubsection{Iterative Self-reinforcement}
\label{self-reinforcement}
The generation paths determined by the above heuristics are model-agnostic and could be noisy and sub-optimal \citep{welleck2019selfreinforcement}. To further improve performance, we allow the model to adjust its own generation paths based on the capabilities it has acquired. That is, transitioning from imitating the oracles to reinforcing its own preferences. In particular, we propose a bootstrapping algorithm to iteratively adjust the target phrases. For each prefix $p$, we first let the model retrieve the $k$-best phrases in the entire candidate pool using its current policy. Then, we choose the valid phrase with the highest semantic matching score from these $k$ phrases as the new target. If no such phrase is found, \textit{i.e.}, none of the \textit{k}-best phrases match the ground-truth continuation, we retain the previous target. The above process is repeated periodically. We present an example in Appendix \ref{self-reinfo examples}.

\vspace{-2pt}
\subsection{Training Objectives}

We optimize our model using the InfoNCE loss \citep{oord2018representation,karpukhin2020dense}, for which a negative phrase set $\mathcal{N}(p)$ is introduced for each triplet $(p, f, s)$.
\begin{gather}
    \label{phrase_loss}
    \mathcal{L}_p = \frac{\exp({E_p(p)\cdot E_c(s)})}{\exp({E_p(p)\cdot E_c(s)}) + \sum_{t \in N(p)} \exp({E_p(p)\cdot E_c(t)})}
\end{gather}
The construction of the negative phrase set $\mathcal{N}(p)$ is detailed below. To preserve the ability for token-level generation, we also train our model with the standard next-token prediction loss $\mathcal{L}_t$ \citep{lan2023cog}. The training objective is formulated as $\mathcal{L}_p + \alpha \mathcal{L}_t$.
\paragraph{Negative Sampling.}
We incorporate two types of negative examples to improve the model's ability to differentiate phrases:
(1) In-batch negatives: We regard all other candidate phrases in the same training batch as this type of negative example. These negatives help the model learn more discriminative representations on a large scale without incurring considerable costs.
(2) Hard negatives: Recall that in Section \ref{self-reinforcement}, we periodically update the generation targets by retrieving top-$k$ candidate phrases for each prefix. Among these $k$ phrases, despite one may be chosen as the new generation target, the remaining phrases can serve as strong negatives because they are likely to confuse the model.

Note that the above negatives may contain false negatives, which are not chosen as targets but still make a valid follow-up. To minimize the risk, we remove all phrases that constitute a prefix of the groundtruth continuation.

\subsection{Models}
\paragraph{Prefix Encoder.}
We treat the prefix as a sequence of tokens with previously predicted
phrases split into tokens. This token sequence is encoded using the standard Transformer architecture with causal attention \citep{vaswani2017Transformers,radford2019gpt2}. The prefix representation is obtained through a linear projection of the last-layer representation of the final token in the sequence.
\paragraph{Phrase Encoder.}
We employ a deep bidirectional Transformer \citep{vaswani2017Transformers,devlin2018bert} to generate contextualized token representations of a supporting document. The representation of a phrase is obtained by concatenating the representations of its first and last tokens, followed by projecting the concatenated representation to the same dimension as the prefix representation. To preserve the ability to compose output using single tokens, we also add the token vocabulary to our phrase table. These standalone tokens can be considered as special phrases, and their representations are obtained through the standard embedding layer of the LM.

\section{Experiment Setup}
\subsection{Implementation Details}
We train our model on the training set of MiniPile\footnote{\url{https://huggingface.co/datasets/JeanKaddour/minipile}}\citep{kaddour2023minipile}, and use the English Wikipedia dump March 1, 2022\footnote{\url{https://huggingface.co/datasets/wikipedia}} as supporting documents. Specifically, we split each Wikipedia article into multiple, disjoint text blocks of up to 128 words as documents, which results in 29,488,431 documents. The size of our phrase index is 137,101,097. We use GPT-2 \citep{radford2019gpt2} and DensePhrases\footnote{\url{https://huggingface.co/princeton-nlp/densephrases-multi}} \citep{lee2021densephrase} to initialize the prefix encoder and the phrase encoder, respectively. For efficiency, we solely fine-tune the prefix encoder. This avoid the computational burden of re-computing phrase embeddings associated with updating the phrase encoder. While revising the training oracles via self-reinforcement, we retrieve the top $k=128$ phrases for each prefix. 

\subsection{Inference Details}
During inference, we employ FAISS \citep{johnson2019faiss}, a library for vector similarity search and clustering, for efficient retrieval.

\paragraph{Continuation Generation.}
\label{Continuation Generation}
For text generation, we directly retrieve top-$k$ candidates from the entire phrase table (including both context-aware phrases and standalone tokens). We then apply a \texttt{softmax} function to the matching scores of these candidates, creating a next-phrase probability distribution \citep{shi2024thorough}, and use top-$p$ sampling \citep{holtzman2019nucleus} for selecting the next phrase. In all experiments, we set $k$ to 128 (see the analysis on $k$ in Table \ref{ablation-k} in Appendix \ref{Appx: ablation studies}) and $p$ to 0.95. To control the ratio of phrase retrieval, we filter out phrases with probabilities below a threshold. The threshold is set to $\phi=0.4$ if not otherwise specified.

\paragraph{Likelihood Estimation.}
\label{Text Probabilities Calculation}
To calculate the likelihood of a given text, we approximate the likelihood by summing all possible generation paths. For instance, given the sentence \textit{"The Moon rises"}, the following generation paths may exist: (1) \textit{The$\rightarrow$moon$\rightarrow$rises}; (2) \textit{The moon$\rightarrow$rises}; (3) \textit{The moon rises}. The probability of each path is the product of the probabilities of all phrases (tokens) along that path. For example, the probability of the path (2) is calculated by $p(rises|The \ moon)\cdot p(The \ moon)$. The probabilities of each step are obtained in the same way as we construct the next-phrase probability distribution for continuation generation. Note that the sum of all possible paths can be computed efficiently using dynamic programming with time complexity $O(n^2)$, where $n$ represents the number of tokens in the text.
% \footnote{$n$ is the number of tokens.}.

\subsection{Baselines}
% We compare with the following baselines.

We compare the proposed method with standard LM in the zero-shot setting, also drawing the following state-of-the-art retrieval-augmented methods as baselines:

\noindent\textbf{Base LM} is the standard token-level language model using the Transformer \citep{vaswani2017Transformers} architecture. We fine-tune the pre-trained GPT-2\footnote{\url{https://huggingface.co/gpt2}} \citep{radford2019gpt2}.

%\noindent\textbf{Base LM (tuned)}. The base LM is further trained on MiniPile and Wikipedia.

\noindent\textbf{$k$NN-LM} \citep{khandelwal2019generalization} is a retrieval-augmented LM that interpolates the next-token distribution of the base LM with a $k$-nearest neighbors ($k$NN) model.

\noindent\textbf{RETRO} \citep{borgeaud2022retro}\footnote{\url{https://github.com/lucidrains/RETRO-pytorch}} is a retrieval-augmented LM incorporated with a pre-trained document retriever, a document encoder and a cross-attention mechanism.%Given the absence of a pre-trained RETRO model, we train it from scratch on our datasets.

\noindent\textbf{CoG} \citep{lan2023cog}\footnote{\url{https://github.com/gmftbyGMFTBY/Copyisallyouneed}} is another retrieval-augmented LM that adopts a two-stage search pipeline. It first retrieves semantically-relevant documents, and then considers all $n$-grams within them as candidate phrases.

\section{Experiments}
We verify the effectiveness of our methods on a set of knowledge-intensive tasks and open-ended text generation tasks without fine-tuning.
\subsection{Knowledge-Intensive Tasks}
\label{classification}
\subsubsection{Datasets}
We employ five knowledge-insensitive datasets, including three open-domain QA datasets: \textbf{OpenbookQA} \citep{mihaylov2018openbookqa}, \textbf{ARC-Challenge} \citep{clark2018ARC}, and \textbf{TruthfulQA} \citep{lin2021truthfulqa}; and two domain-specific (medical) datasets: \textbf{MedMCQA} \citep{medmcqa} and \textbf{Med-USMILE} \citep{jin202medusmile}. The details for these datasets can be found in Appendix \ref{sec:appendix}.

In line with prior research~\citep{2020BrownGPT3, 2021SanhT0}, we adopt a \textit{classification with options} methodology to quantify the model performance. This approach involves presenting the model with a range of options and calculating the likelihood of each option being the correct response. The option with the highest probability is selected as the model's prediction. We then report the \textbf{accuracy} of the model's predictions.

\begin{table*}[t]
\normalsize
\centering
\setlength{\tabcolsep}{2.4pt}
\begin{tabular}{cccccc}
\toprule
& \textbf{TruthfulQA} & \textbf{OpenbookQA} & \textbf{ARC-Challenge} & \textbf{MedMCQA} & \textbf{Med-USMILE} \\
\hline
%random & 22.61 & 25.00 & 25.07 & 25.00 & 25.00 \\
Base LM (w/o FT)& 30.27 & 22.67 & 24.52 & 27.96 & 24.89 \\
Base LM  & 29.73 & 23.47 & 23.92 & 28.33 & 24.19 \\
$k$NN-LM & 30.27 & 22.93 & 24.82 & 27.96 & 24.72 \\
RETRO & 27.53 & 26.13 & 22.21 & 25.68 & 25.33 \\
CoG & 34.11 & 35.47 & 27.24 & 29.07 & 25.07\\
Ours & \textbf{34.27} & \textbf{36.27} & \textbf{28.27} & \textbf{29.44} & \textbf{25.69} \\
Ours (w/o phrase) & 28.63 & 23.73 & 22.51 & 27.42 & 24.80 \\
\bottomrule
\end{tabular}
\caption{\label{from pre-trained result}
Experiments on knowledge-intensive tasks. Ours (w/o phrase): a variant of our model that restricts the model to only use standalone tokens without retrieving context-aware phrases.}
\end{table*}

\subsubsection{Results}
We compare our methods with baselines in knowledge-intensive tasks across several settings.
\paragraph{Main Results.}
As shown in Table \ref{from pre-trained result}, our model consistently outperforms various baseline models across all datasets. Compared with base LM, our model improves the accuracy of the TruthfulQA and OpenBookQA datasets from 29.73\% to 34.27\% and 23.47\% to 36.27\%, respectively.
When we eliminate the phrase retrieval from our model and only use standalone tokens (Ours w/o phrase), there is a considerable drop in performance, demonstrating the effectiveness of incorporating phrase retrieval in our methods. Note that the models presented in Table \ref{from pre-trained result} are initialized from pre-trained LMs. To analyze the role of pre-trained models in our framework, we train all models from scratch with random initialization. The results are shown in Table \ref{from scratch result} in Appendix \ref{Appx: ablation studies}, our model outperforms the baselines across all datasets. For example, our model achieves a 12.8\% absolute improvement on OpenbookQA over base LM, suggesting that our training framework is not heavily dependent on pre-trained models. To elucidate the role of phrase retrieval in knowledge-intensive tasks, we delve into a case study depicted in Appendix \ref{Appx: case study}.

% \begin{table*}[t]
% \normalsize
% \centering
% \setlength{\tabcolsep}{2.4pt}
% \begin{tabular}{cccccc}
% \toprule
% & \textbf{TruthfulQA} & \textbf{OpenbookQA} & \textbf{ARC-Challenge} & \textbf{MedMCQA} & \textbf{Med-USMILE} \\
% \hline
% %random & 22.61 & 25.00 & 25.07 & 25.00 & 25.00 \\
% Base LM (w/o FT)& 30.27 & 22.67 & 24.52 & 27.96 & 24.89 \\
% Base LM  & 29.73 & 23.47 & 23.92 & 28.33 & 24.19 \\
% $k$NN-LM & 30.27 & 22.93 & 24.82 & 27.96 & 24.72 \\
% RETRO & 27.53 & 26.13 & 22.21 & 25.68 & 25.33 \\
% CoG & 34.11 & 35.47 & 27.24 & 29.07 & 25.07\\
% \hline
% \noalign{\smallskip}
% Ours & 34.27 & 36.27 & \textbf{28.27} & 29.44 & 25.69 \\
% w/o phrase & 28.63 & 23.73 & 22.51 & 27.42 & 24.80 \\
% w/ enlarged index & \textbf{39.59} & \textbf{37.07} & 27.14 & \textbf{31.63} & \textbf{27.87} \\
% \bottomrule
% \end{tabular}
% \caption{\label{from pre-trained result}
% Experiments on knowledge-intensive tasks.
% % We report the accuracy on all tasks.
% % random: random choice. 
% Ours (w/o phrase): a variant of our model that restricts the model to only use standalone tokens without retrieving context-aware phrases.}
% \end{table*}

\begin{table*}[t]
\normalsize
\centering
\setlength{\tabcolsep}{2.5pt}
\begin{tabular}{cccccc}
\toprule
& \textbf{TruthfulQA} & \textbf{OpenbookQA} & \textbf{ARC-Challenge} & \textbf{MedMCQA} & \textbf{Med-USMILE} \\
\hline
Ours & 34.27 & 36.27 & \textbf{28.27} & 29.44 & 25.69 \\
w/ enlarged index & \textbf{39.59} & \textbf{37.07} & 27.14 & \textbf{31.63} & \textbf{27.87} \\
\bottomrule
\end{tabular}
\caption{\label{enlarged index}
Results for our model with an enlarged phrase index.}
\end{table*}

\paragraph{Enlarged Phrase Index.}
Recall that we exclude phrases with excessively high or low IDF values (Section \ref{pseudo label}). This strategy not only stabilizes the training process but also improves training efficiency. However, the phrases initially filtered out can be repurposed to augment our phrase index in a training-free manner. 
This expanded phrase index, now three times larger than the original, underscores the scalability of our approach. 
As evidenced in Table \ref{enlarged index}, this expansion boosts our model's performance, such as a 5.32\% increase in accuracy on TruthfulQA.  
This not only highlights our model's potential to generalize to unseen phrases and documents but also emphasizes its plug-and-play feature, capable of adapting to a larger phrase table without the need for re-training.

\begin{wraptable}{r}{0.52\textwidth}
  \vspace{-3.5mm}
  \centering
  \begin{tabular}{ccc}
    \hline
     & \textbf{MedMCQA} & \textbf{Med-USMILE} \\
    \hline
    Base LM (FT) & 28.79 & 25.15 \\
    General index & 29.44 & 25.69 \\
    Medical index & \textbf{29.50} & \textbf{26.38} \\
    w/o phrase & 27.42 & 24.80 \\
    \bottomrule
  \end{tabular}
  \caption{Results on medical datasets.}
  \label{medical index}
\end{wraptable}

\paragraph{Domain Adaption.}
The \textit{plug-and-play} property of the phrase index further motivates us to employ a domain-specific index for the QA tasks in the medical domain without any domain-specific training.
To this end, we construct an index consisting of 3 million phrases by extracting phrases from a small text collection of the medical domain\footnote{\url{https://huggingface.co/datasets/gamino/wiki\_medical\_terms}}. For comparison purpose, we also fine-tune the base LM on it for fair comparison. As illustrated in Table \ref{medical index}, despite the considerable reduction in index size compared to the original Wikipedia index (3 million vs 137 million), our model exhibits even better performance on two medical QA datasets. This result underscores our model's capability to enhance its performance in specific domains by leveraging a domain-specific, well-curated phrase index in a training-free manner.

\subsection{Open-Ended Text Generation}
\label{generation}

We conduct open-ended text generation experiments on the test set of MiniPile \citep{kaddour2023minipile}. For each document in the test set, we adopt the first 128 tokens as the prefix. The baselines and our model are required to generate text continuations of 128 tokens in length based on the same prefix. 
% Given a text prefix, our model repeatedly retrieves the top-$k$ phrases from the combined set of all candidate phrases and tokens, and then selects one candidate as the continuation based on the probability distribution that is calculated among these $k$ candidates by applying Softmax.

\subsubsection{Evaluation Metrics}
Following previous works \citep{welleck2019metric1,su2022metric2,lan2023cog}, we utilize three automatic evaluation metrics to measure the quality of the generated texts: 
(i) \textbf{MAUVE} \citep{pillutla2021mauve} captures the overall usefulness of the generated text by estimating the average utility of the content; (ii) \textbf{Coherence} measures the logical consistency and flow of the generated text, ensuring that the output is well-structured and easy to understand; and (iii) \textbf{Diversity} evaluates the variety of generated content, promoting the generation of unique and creative text. We report MAUVE and diversity as percentages (\%). The details for these metrics can be found in Appendix \ref{Appx: metrics}.
We also measure the average time cost for a model to decode a continuation consisting of 128 tokens given a prefix of 128 tokens, referred to as \textbf{latency}. 

\begin{table}[t]
\normalsize
\centering
\setlength{\tabcolsep}{11pt}
\begin{tabular}{ccccc}
\toprule
& \textbf{MAUVE}$\uparrow$ & \textbf{Coherence}$\downarrow$ & \textbf{Diversity}$\uparrow$ & \textbf{Latency}$\downarrow$ \\
\hline
Base LM (w/o FT) & 69.68 & 3.64 & 83.14 & 1.00x\\
Base LM & 42.61 & 3.56 & 78.72 & 1.00x \\
$k$NN-LM & 13.07 & 5.63 & \textbf{88.10} & 6.29x \\
RETRO & 62.39 & 4.82 & 80.96 & 1.51x \\
CoG & 52.27 & \textbf{2.08} & 55.04 & 4.40x \\
Ours & \textbf{81.58} & 3.25 & 76.26 & \textbf{1.29x} \\
\bottomrule
\end{tabular}
\caption{\label{open-ended generation}
Results for open-ended text generation.}
\end{table}

\begin{table}[t]
\normalsize
\centering
\setlength{\tabcolsep}{10pt}
\begin{tabular}{ccccc}
\toprule
\textbf{Model} & \textbf{Fluency} & \textbf{Coherence} & \textbf{Informativeness} & \textbf{Grammar} \\
\hline
Base LM (w/o FT)& 2.91 & 2.33 & 2.35 & 3.00 \\
Base LM & 2.81 & 2.37 & 2.40 & 2.79 \\
Ours & \textbf{2.95} & \textbf{2.70} & \textbf{2.67} & \textbf{3.02} \\
\hline
\end{tabular}
\caption{Human evaluation results.}
\label{table:human eval}
\end{table}

\subsubsection{Results}
As shown in Table \ref{open-ended generation}, 
our model attains the highest MAUVE score among all models, demonstrating the high quality of the generated text. Other retrieval-augmented methods underperform base LM in the MAUVE score due to text degeneration, which aligns with findings in previous work \citep{wang2023knnlmdoesnot}.
Our model also shows a strong balance between coherence and diversity.
The coherence score of our model is 3.25, which outperforms most baselines except for CoG. However, we find that CoG often generates lexically similar, meaningless sentences, which is reflected in its low diversity score of 55.04\%.
Meanwhile, our model's diversity score is 76.26\%, which is slightly lower than some baseline models, but these models often generate incoherent sentences, as reflected in their lower coherence scores.

% Moreover, the latency of our model is slightly higher than Base LM but significantly lower than other retrieval-augmented models. We provide a detailed analysis of the computational cost in Section~\ref{sec:cc}.
\paragraph{Human Evaluation.}
To gain further insights, we randomly sample 100 cases and evaluate the results of the base LM, the base LM without fine-tuning (w/o FT), and our model from four perspectives: fluency, coherence, informativeness, and grammar. Each aspect is scored on a Likert scale from 1 to 4 (1 represents \textit{"bad"}, 2 stands for \textit{"fair"}, 3 is considered \textit{"good"}, and 4 signifies \textit{"very good"}). 
We report the average scores in table \ref{table:human eval}. As we can see, our method outperforms the base LM in all four categories, especially in coherence and informativeness. This indicates that our model, based on phrase retrieval, is better at following the preceding context and providing more informative content. As for the lower scores of the base LM compared to the base LM (w/o FT), we find that they are largely due to formatting issues. Further analysis can be found in Appendix \ref{Appx: Detailed Human Evaluation}. 
\paragraph{Generation Speed.}
\label{sec:cc}
We now discuss the generation latency of different models. 
In Table \ref{open-ended generation}, we report the relative latency, taking the base LM as the baseline. 
$k$NN-LM incurs the highest cost due to the need for interpolating the base LM's token distribution with another distribution computed using its datastore.
The CoG model also exhibits a notable overhead as it involves extracting all $n$-grams from the retrieved documents, applying \texttt{softmax} over tokens and all $n$-grams, and sampling from the resulting probability distribution.
The RETRO model, although faster than the previous two, still requires time for applying the representations of retrieved text chunks in attention computation.
Our method stands out with the highest generation speed, since it directly retrieves and utilizes phrases.

% \vspace{-10mm}
\begin{wraptable}{r}{0.43\textwidth}
  \vspace{-3.5mm}
  \centering
  \begin{tabular}{cccc}
    \hline
     & \textbf{MAUVE}$\uparrow$ & \textbf{Coh.}$\downarrow$ & \textbf{Div.}$\uparrow$ \\
    \hline
    w/o SR & 7.86 & 4.14 & {\bf 81.14} \\
    round1 & 64.49 & \textbf{3.23} & 70.15 \\
    round2 & {\bf 81.58} & 3.25 & 76.26 \\
    % round3 & 74.94 & \textbf{3.04} & 73.76 \\
    \hline
  \end{tabular}
  \caption{Ablation study on the effect of self-reinforcement.}
  \label{ablation-SR-generation}
\end{wraptable}

\paragraph{Effect of Self-reinforcement.}
Ablation studies on the effect of the Self-Reinforcement (SR) mechanism reveal significant insights into the performance of our model. In the case of knowledge-intensive tasks, we do not observe a significant impact of SR on our model's performance (refer to Table \ref{ablation-SR-QA} in Appendix \ref{Appx: ablation studies}). This suggests that our framework is inherently effective in handling such tasks, even without the aid of SR.
However, the scenario differs for open-ended text generation. Table \ref{ablation-SR-generation} shows that models trained with SR exhibit substantial improvements in the MAUVE scores across multiple rounds, which indicates the importance of SR in enhancing the quality of text generation.
After the second round, we do not observe noticeable improvements with additional rounds of SR iteration, suggesting that the model converges to its optimal state.

\section{Related Work}
Standard language models (LMs) \citep{radford2019gpt2,2020BrownGPT3} are trained to predict the next token given a text prefix. With a vast amount of training corpora and model parameters, these models show strong zero-shot performance on various downstream tasks, serving as a unified solution for natural language processing. However, scaling up the model parameters and training corpora can be very expensive and cannot be done in a timely manner.

To tackle the above issues, there has been an increasing body of work that enhances the parametric LM with a non-parametric component \citep{Li2022ASO}. \citet{guu2020realm,lewis2020retrieval,borgeaud2022retro,izacard2022few} ground the next token prediction on a set of relevant documents obtained using retrieval techniques \citep{Robertson2009ThePR,karpukhin2020dense}. \citet{khandelwal2019generalization,yogatama2021adaptive,zhong2022training}
augment the output probability distribution with non-parametric nearest neighborhood estimation. Also, the retrieve-then-generate paradigm has been extensively studied in specific downstream tasks, such as code generation \citep{hashimoto2018retrieve}, question answering \citep{ye2023fits,karpukhin2020dense,lee2021learning}, open-domain dialogue systems \citep{Weston2018RetrieveAR,wu2019response,cai2019skeleton,cai-etal-2019-retrieval}, and machine translation \citep{khandelwal2021nearest,cai-etal-2021-neural}, multimodal retrieval \citep{jin2023diffusionret,li2023freestyleret}.

The work most closely related to ours is that of \citet{min2022nonparametric} and \citet{lan2023cog}. The former explores a similar idea in the area of masked language models to enhance natural language understanding. \citet{lan2023cog}, on the other hand, allows the copy of phrases from the grounding documents. However, their approach still relies on a two-stage pipeline, grounding the generation on a small set of retrieved documents only. While \citet{lan2023cog} simply employs the longest common subsequence algorithm to find phrases that can be copied from the retrieved documents, we present heuristics-based and self-reinforced mechanisms to construct reliable training oracles. Also, \citet{lan2023cog} only evaluates the performance on open-ended text generation tasks.

\section{Conclusion}

We presented a novel retrieval-based text generation approach using context-aware phrase retrieval. Our method addresses the primary challenge of constructing training oracles through heuristic-based initialization and iterative self-reinforcement. 
Experiments on knowledge-intensive tasks and open-ended text generation tasks show that the proposed method outperforms the standard LM and state-of-the-art retrieval-augmented methods.
Moreover, our model exhibits superior performance with either an enlarged or a smaller, domain-specific index, and achieves the lowest generation latency compared to other retrieval-augmented baselines.
This work contributes to the NLP research community by promoting a paradigm shift towards more accurate generation via retrieval.
As we continue to explore and refine the paradigm, we invite readers to consider the limitations of our current work, as detailed in Appendix \ref{limitation}, to fully appreciate the scope of future research.

\bibliography{iclr2024_conference}

\begin{thebibliography}{63}
\providecommand{\natexlab}[1]{#1}
\providecommand{\url}[1]{\texttt{#1}}
\expandafter\ifx\csname urlstyle\endcsname\relax
  \providecommand{\doi}[1]{doi: #1}\else
  \providecommand{\doi}{doi: \begingroup \urlstyle{rm}\Url}\fi

\bibitem[Asai et~al.(2023)Asai, Min, Zhong, and Chen]{asai2023retrieval}
Akari Asai, Sewon Min, Zexuan Zhong, and Danqi Chen.
\newblock Retrieval-based language models and applications.
\newblock In \emph{Proceedings of the 61st Annual Meeting of the Association
  for Computational Linguistics (Volume 6: Tutorial Abstracts)}, 2023.

\bibitem[Borgeaud et~al.(2022)Borgeaud, Mensch, Hoffmann, Cai, Rutherford,
  Millican, van~den Driessche, Lespiau, Damoc, Clark, de~Las~Casas, Guy,
  Menick, Ring, Hennigan, Huang, Maggiore, Jones, Cassirer, Brock, Paganini,
  Irving, Vinyals, Osindero, Simonyan, Rae, Elsen, and
  Sifre]{borgeaud2022retro}
Sebastian Borgeaud, Arthur Mensch, Jordan Hoffmann, Trevor Cai, Eliza
  Rutherford, Katie Millican, George van~den Driessche, Jean{-}Baptiste
  Lespiau, Bogdan Damoc, Aidan Clark, Diego de~Las~Casas, Aurelia Guy, Jacob
  Menick, Roman Ring, Tom Hennigan, Saffron Huang, Loren Maggiore, Chris Jones,
  Albin Cassirer, Andy Brock, Michela Paganini, Geoffrey Irving, Oriol Vinyals,
  Simon Osindero, Karen Simonyan, Jack~W. Rae, Erich Elsen, and Laurent Sifre.
\newblock Improving language models by retrieving from trillions of tokens.
\newblock In Kamalika Chaudhuri, Stefanie Jegelka, Le~Song, Csaba
  Szepesv{\'{a}}ri, Gang Niu, and Sivan Sabato (eds.), \emph{International
  Conference on Machine Learning, {ICML} 2022, 17-23 July 2022, Baltimore,
  Maryland, {USA}}, volume 162 of \emph{Proceedings of Machine Learning
  Research}, 2022.

\bibitem[Brown et~al.(2020)Brown, Mann, Ryder, Subbiah, Kaplan, Dhariwal,
  Neelakantan, Shyam, Sastry, Askell, Agarwal, Herbert{-}Voss, Krueger,
  Henighan, Child, Ramesh, Ziegler, Wu, Winter, Hesse, Chen, Sigler, Litwin,
  Gray, Chess, Clark, Berner, McCandlish, Radford, Sutskever, and
  Amodei]{2020BrownGPT3}
Tom~B. Brown, Benjamin Mann, Nick Ryder, Melanie Subbiah, Jared Kaplan,
  Prafulla Dhariwal, Arvind Neelakantan, Pranav Shyam, Girish Sastry, Amanda
  Askell, Sandhini Agarwal, Ariel Herbert{-}Voss, Gretchen Krueger, Tom
  Henighan, Rewon Child, Aditya Ramesh, Daniel~M. Ziegler, Jeffrey Wu, Clemens
  Winter, Christopher Hesse, Mark Chen, Eric Sigler, Mateusz Litwin, Scott
  Gray, Benjamin Chess, Jack Clark, Christopher Berner, Sam McCandlish, Alec
  Radford, Ilya Sutskever, and Dario Amodei.
\newblock Language models are few-shot learners.
\newblock In Hugo Larochelle, Marc'Aurelio Ranzato, Raia Hadsell,
  Maria{-}Florina Balcan, and Hsuan{-}Tien Lin (eds.), \emph{Advances in Neural
  Information Processing Systems 33: Annual Conference on Neural Information
  Processing Systems 2020, NeurIPS 2020, December 6-12, 2020, virtual}, 2020.

\bibitem[Cai et~al.(2019{\natexlab{a}})Cai, Wang, Bi, Tu, Liu, Lam, and
  Shi]{cai2019skeleton}
Deng Cai, Yan Wang, Wei Bi, Zhaopeng Tu, Xiaojiang Liu, Wai Lam, and Shuming
  Shi.
\newblock Skeleton-to-response: Dialogue generation guided by retrieval memory.
\newblock In \emph{Proceedings of the 2019 Conference of the North {A}merican
  Chapter of the Association for Computational Linguistics: Human Language
  Technologies, Volume 1 (Long and Short Papers)}, 2019{\natexlab{a}}.

\bibitem[Cai et~al.(2019{\natexlab{b}})Cai, Wang, Bi, Tu, Liu, and
  Shi]{cai-etal-2019-retrieval}
Deng Cai, Yan Wang, Wei Bi, Zhaopeng Tu, Xiaojiang Liu, and Shuming Shi.
\newblock Retrieval-guided dialogue response generation via a
  matching-to-generation framework.
\newblock In \emph{Proceedings of the 2019 Conference on Empirical Methods in
  Natural Language Processing and the 9th International Joint Conference on
  Natural Language Processing (EMNLP-IJCNLP)}, 2019{\natexlab{b}}.

\bibitem[Cai et~al.(2021)Cai, Wang, Li, Lam, and Liu]{cai-etal-2021-neural}
Deng Cai, Yan Wang, Huayang Li, Wai Lam, and Lemao Liu.
\newblock Neural machine translation with monolingual translation memory.
\newblock In \emph{Proceedings of the 59th Annual Meeting of the Association
  for Computational Linguistics and the 11th International Joint Conference on
  Natural Language Processing (Volume 1: Long Papers)}, 2021.

\bibitem[Chomsky(1957)]{chomsky1957syntactic}
Noam Chomsky.
\newblock Syntactic structures.
\newblock 1957.

\bibitem[Clark et~al.(2018)Clark, Cowhey, Etzioni, Khot, Sabharwal, Schoenick,
  and Tafjord]{clark2018ARC}
Peter Clark, Isaac Cowhey, Oren Etzioni, Tushar Khot, Ashish Sabharwal, Carissa
  Schoenick, and Oyvind Tafjord.
\newblock Think you have solved question answering? try arc, the ai2 reasoning
  challenge.
\newblock \emph{ArXiv preprint}, abs/1803.05457, 2018.

\bibitem[Cruse(1986)]{cruse1986polysemy}
D~Alan Cruse.
\newblock \emph{Lexical semantics}.
\newblock 1986.

\bibitem[Devlin et~al.(2019)Devlin, Chang, Lee, and Toutanova]{devlin2018bert}
Jacob Devlin, Ming-Wei Chang, Kenton Lee, and Kristina Toutanova.
\newblock {BERT}: Pre-training of deep bidirectional transformers for language
  understanding.
\newblock In \emph{Proceedings of the 2019 Conference of the North {A}merican
  Chapter of the Association for Computational Linguistics: Human Language
  Technologies, Volume 1 (Long and Short Papers)}, 2019.

\bibitem[Dyer et~al.(2016)Dyer, Kuncoro, Ballesteros, and
  Smith]{dyer-etal-2016-recurrent}
Chris Dyer, Adhiguna Kuncoro, Miguel Ballesteros, and Noah~A. Smith.
\newblock Recurrent neural network grammars.
\newblock In \emph{Proceedings of the 2016 Conference of the North {A}merican
  Chapter of the Association for Computational Linguistics: Human Language
  Technologies}, 2016.

\bibitem[Guo et~al.(2016)Guo, Kumar, Choromanski, and Simcha]{guo2016mips2}
Ruiqi Guo, Sanjiv Kumar, Krzysztof Choromanski, and David Simcha.
\newblock Quantization based fast inner product search.
\newblock In Arthur Gretton and Christian~C. Robert (eds.), \emph{Proceedings
  of the 19th International Conference on Artificial Intelligence and
  Statistics, {AISTATS} 2016, Cadiz, Spain, May 9-11, 2016}, volume~51 of
  \emph{{JMLR} Workshop and Conference Proceedings}, 2016.

\bibitem[Guu et~al.(2020)Guu, Lee, Tung, Pasupat, and Chang]{guu2020realm}
Kelvin Guu, Kenton Lee, Zora Tung, Panupong Pasupat, and Ming-Wei Chang.
\newblock Realm: Retrieval-augmented language model pre-training.
\newblock \emph{ArXiv preprint}, abs/2002.08909, 2020.

\bibitem[Hashimoto et~al.(2018)Hashimoto, Guu, Oren, and
  Liang]{hashimoto2018retrieve}
Tatsunori~B. Hashimoto, Kelvin Guu, Yonatan Oren, and Percy Liang.
\newblock A retrieve-and-edit framework for predicting structured outputs.
\newblock In Samy Bengio, Hanna~M. Wallach, Hugo Larochelle, Kristen Grauman,
  Nicol{\`{o}} Cesa{-}Bianchi, and Roman Garnett (eds.), \emph{Advances in
  Neural Information Processing Systems 31: Annual Conference on Neural
  Information Processing Systems 2018, NeurIPS 2018, December 3-8, 2018,
  Montr{\'{e}}al, Canada}, 2018.

\bibitem[Holtzman et~al.(2020)Holtzman, Buys, Du, Forbes, and
  Choi]{holtzman2019nucleus}
Ari Holtzman, Jan Buys, Li~Du, Maxwell Forbes, and Yejin Choi.
\newblock The curious case of neural text degeneration.
\newblock In \emph{8th International Conference on Learning Representations,
  {ICLR} 2020, Addis Ababa, Ethiopia, April 26-30, 2020}, 2020.

\bibitem[Izacard et~al.(2022)Izacard, Lewis, Lomeli, Hosseini, Petroni, Schick,
  Dwivedi-Yu, Joulin, Riedel, and Grave]{izacard2022few}
Gautier Izacard, Patrick Lewis, Maria Lomeli, Lucas Hosseini, Fabio Petroni,
  Timo Schick, Jane Dwivedi-Yu, Armand Joulin, Sebastian Riedel, and Edouard
  Grave.
\newblock Few-shot learning with retrieval augmented language models.
\newblock \emph{ArXiv preprint}, abs/2208.03299, 2022.

\bibitem[Jin et~al.(2021)Jin, Pan, Oufattole, Weng, Fang, and
  Szolovits]{jin202medusmile}
Di~Jin, Eileen Pan, Nassim Oufattole, Wei-Hung Weng, Hanyi Fang, and Peter
  Szolovits.
\newblock What disease does this patient have? a large-scale open domain
  question answering dataset from medical exams.
\newblock \emph{Applied Sciences}, 11\penalty0 (14), 2021.

\bibitem[Jin et~al.(2023)Jin, Li, Cheng, Li, Ji, Liu, Yuan, and
  Chen]{jin2023diffusionret}
Peng Jin, Hao Li, Zesen Cheng, Kehan Li, Xiangyang Ji, Chang Liu, Li~Yuan, and
  Jie Chen.
\newblock Diffusionret: Generative text-video retrieval with diffusion model.
\newblock \emph{arXiv preprint arXiv:2303.09867}, 2023.

\bibitem[Johnson et~al.(2019)Johnson, Douze, and J{\'e}gou]{johnson2019faiss}
Jeff Johnson, Matthijs Douze, and Herv{\'e} J{\'e}gou.
\newblock Billion-scale similarity search with gpus.
\newblock \emph{IEEE Transactions on Big Data}, 7\penalty0 (3), 2019.

\bibitem[Kaddour(2023)]{kaddour2023minipile}
Jean Kaddour.
\newblock The minipile challenge for data-efficient language models, 2023.

\bibitem[Karpukhin et~al.(2020)Karpukhin, Oguz, Min, Lewis, Wu, Edunov, Chen,
  and Yih]{karpukhin2020dense}
Vladimir Karpukhin, Barlas Oguz, Sewon Min, Patrick Lewis, Ledell Wu, Sergey
  Edunov, Danqi Chen, and Wen-tau Yih.
\newblock Dense passage retrieval for open-domain question answering.
\newblock In \emph{Proceedings of the 2020 Conference on Empirical Methods in
  Natural Language Processing (EMNLP)}, 2020.

\bibitem[Khandelwal et~al.(2020)Khandelwal, Levy, Jurafsky, Zettlemoyer, and
  Lewis]{khandelwal2019generalization}
Urvashi Khandelwal, Omer Levy, Dan Jurafsky, Luke Zettlemoyer, and Mike Lewis.
\newblock Generalization through memorization: Nearest neighbor language
  models.
\newblock In \emph{8th International Conference on Learning Representations,
  {ICLR} 2020, Addis Ababa, Ethiopia, April 26-30, 2020}, 2020.

\bibitem[Khandelwal et~al.(2021)Khandelwal, Fan, Jurafsky, Zettlemoyer, and
  Lewis]{khandelwal2021nearest}
Urvashi Khandelwal, Angela Fan, Dan Jurafsky, Luke Zettlemoyer, and Mike Lewis.
\newblock Nearest neighbor machine translation.
\newblock In \emph{9th International Conference on Learning Representations,
  {ICLR} 2021, Virtual Event, Austria, May 3-7, 2021}, 2021.

\bibitem[Lan et~al.(2023)Lan, Cai, Wang, Huang, and Mao]{lan2023cog}
Tian Lan, Deng Cai, Yan Wang, Heyan Huang, and Xian-Ling Mao.
\newblock Copy is all you need.
\newblock In \emph{The Eleventh International Conference on Learning
  Representations}, 2023.

\bibitem[Lee et~al.(2021{\natexlab{a}})Lee, Sung, Kang, and
  Chen]{lee2021learning}
Jinhyuk Lee, Mujeen Sung, Jaewoo Kang, and Danqi Chen.
\newblock Learning dense representations of phrases at scale.
\newblock In \emph{Proceedings of the 59th Annual Meeting of the Association
  for Computational Linguistics and the 11th International Joint Conference on
  Natural Language Processing (Volume 1: Long Papers)}, 2021{\natexlab{a}}.

\bibitem[Lee et~al.(2021{\natexlab{b}})Lee, Wettig, and
  Chen]{lee2021densephrase}
Jinhyuk Lee, Alexander Wettig, and Danqi Chen.
\newblock Phrase retrieval learns passage retrieval, too.
\newblock In \emph{Proceedings of the 2021 Conference on Empirical Methods in
  Natural Language Processing}, 2021{\natexlab{b}}.

\bibitem[Lewis et~al.(2020)Lewis, Perez, Piktus, Petroni, Karpukhin, Goyal,
  K{\"{u}}ttler, Lewis, Yih, Rockt{\"{a}}schel, Riedel, and
  Kiela]{lewis2020retrieval}
Patrick S.~H. Lewis, Ethan Perez, Aleksandra Piktus, Fabio Petroni, Vladimir
  Karpukhin, Naman Goyal, Heinrich K{\"{u}}ttler, Mike Lewis, Wen{-}tau Yih,
  Tim Rockt{\"{a}}schel, Sebastian Riedel, and Douwe Kiela.
\newblock Retrieval-augmented generation for knowledge-intensive {NLP} tasks.
\newblock In Hugo Larochelle, Marc'Aurelio Ranzato, Raia Hadsell,
  Maria{-}Florina Balcan, and Hsuan{-}Tien Lin (eds.), \emph{Advances in Neural
  Information Processing Systems 33: Annual Conference on Neural Information
  Processing Systems 2020, NeurIPS 2020, December 6-12, 2020, virtual}, 2020.

\bibitem[Li et~al.(2023{\natexlab{a}})Li, Jia, Jin, Cheng, Li, Sui, Liu, and
  Yuan]{li2023freestyleret}
Hao Li, Curise Jia, Peng Jin, Zesen Cheng, Kehan Li, Jialu Sui, Chang Liu, and
  Li~Yuan.
\newblock Freestyleret: Retrieving images from style-diversified queries.
\newblock \emph{arXiv preprint arXiv:2312.02428}, 2023{\natexlab{a}}.

\bibitem[Li et~al.(2022)Li, Su, Cai, Wang, and Liu]{Li2022ASO}
Huayang Li, Yixuan Su, Deng Cai, Yan Wang, and Lemao Liu.
\newblock A survey on retrieval-augmented text generation.
\newblock \emph{ArXiv preprint}, abs/2202.01110, 2022.

\bibitem[Li et~al.(2023{\natexlab{b}})Li, Cui, Yan, Yin, Bi, Shi, and
  Zhang]{li-etal-2023-explicit}
Yafu Li, Leyang Cui, Jianhao Yan, Yongjing Yin, Wei Bi, Shuming Shi, and Yue
  Zhang.
\newblock Explicit syntactic guidance for neural text generation.
\newblock In \emph{Proceedings of the 61st Annual Meeting of the Association
  for Computational Linguistics (Volume 1: Long Papers)}, 2023{\natexlab{b}}.

\bibitem[Lin et~al.(2022)Lin, Hilton, and Evans]{lin2021truthfulqa}
Stephanie Lin, Jacob Hilton, and Owain Evans.
\newblock {T}ruthful{QA}: Measuring how models mimic human falsehoods.
\newblock In \emph{Proceedings of the 60th Annual Meeting of the Association
  for Computational Linguistics (Volume 1: Long Papers)}, 2022.

\bibitem[Mihaylov et~al.(2018)Mihaylov, Clark, Khot, and
  Sabharwal]{mihaylov2018openbookqa}
Todor Mihaylov, Peter Clark, Tushar Khot, and Ashish Sabharwal.
\newblock Can a suit of armor conduct electricity? a new dataset for open book
  question answering.
\newblock In \emph{Proceedings of the 2018 Conference on Empirical Methods in
  Natural Language Processing}, 2018.

\bibitem[Mikolov et~al.(2010)Mikolov, Karafi{\'a}t, Burget, Cernock{\`y}, and
  Khudanpur]{mikolov2010recurrent}
Tomas Mikolov, Martin Karafi{\'a}t, Lukas Burget, Jan Cernock{\`y}, and Sanjeev
  Khudanpur.
\newblock Recurrent neural network based language model.
\newblock In \emph{Interspeech}, volume~2. Makuhari, 2010.

\bibitem[Mikolov et~al.(2013)Mikolov, Sutskever, Chen, Corrado, and
  Dean]{mikolov2013neg1}
Tom{\'{a}}s Mikolov, Ilya Sutskever, Kai Chen, Gregory~S. Corrado, and Jeffrey
  Dean.
\newblock Distributed representations of words and phrases and their
  compositionality.
\newblock In Christopher J.~C. Burges, L{\'{e}}on Bottou, Zoubin Ghahramani,
  and Kilian~Q. Weinberger (eds.), \emph{Advances in Neural Information
  Processing Systems 26: 27th Annual Conference on Neural Information
  Processing Systems 2013. Proceedings of a meeting held December 5-8, 2013,
  Lake Tahoe, Nevada, United States}, 2013.

\bibitem[Min et~al.(2019)Min, Chen, Hajishirzi, and
  Zettlemoyer]{min2019discrete}
Sewon Min, Danqi Chen, Hannaneh Hajishirzi, and Luke Zettlemoyer.
\newblock A discrete hard {EM} approach for weakly supervised question
  answering.
\newblock In \emph{Proceedings of the 2019 Conference on Empirical Methods in
  Natural Language Processing and the 9th International Joint Conference on
  Natural Language Processing (EMNLP-IJCNLP)}, 2019.

\bibitem[Min et~al.(2022)Min, Shi, Lewis, Chen, Yih, Hajishirzi, and
  Zettlemoyer]{min2022nonparametric}
Sewon Min, Weijia Shi, Mike Lewis, Xilun Chen, Wen-tau Yih, Hannaneh
  Hajishirzi, and Luke Zettlemoyer.
\newblock Nonparametric masked language modeling.
\newblock \emph{ArXiv preprint}, abs/2212.01349, 2022.

\bibitem[Morgan \& Newport(1981)Morgan and Newport]{morgan1981role}
James~L Morgan and Elissa~L Newport.
\newblock The role of constituent structure in the induction of an artificial
  language.
\newblock \emph{Journal of verbal learning and verbal behavior}, 20\penalty0
  (1), 1981.

\bibitem[Oord et~al.(2018)Oord, Li, and Vinyals]{oord2018representation}
Aaron van~den Oord, Yazhe Li, and Oriol Vinyals.
\newblock Representation learning with contrastive predictive coding.
\newblock \emph{ArXiv preprint}, abs/1807.03748, 2018.

\bibitem[OpenAI(2022)]{chatgpt}
OpenAI.
\newblock Introducing chatgpt.
\newblock 2022.

\bibitem[Pal et~al.(2022)Pal, Umapathi, and Sankarasubbu]{medmcqa}
Ankit Pal, Logesh~Kumar Umapathi, and Malaikannan Sankarasubbu.
\newblock Medmcqa: A large-scale multi-subject multi-choice dataset for medical
  domain question answering.
\newblock In Gerardo Flores, George~H Chen, Tom Pollard, Joyce~C Ho, and
  Tristan Naumann (eds.), \emph{Proceedings of the Conference on Health,
  Inference, and Learning}, volume 174 of \emph{Proceedings of Machine Learning
  Research}, 2022.

\bibitem[Pillutla et~al.(2021)Pillutla, Swayamdipta, Zellers, Thickstun,
  Welleck, Choi, and Harchaoui]{pillutla2021mauve}
Krishna Pillutla, Swabha Swayamdipta, Rowan Zellers, John Thickstun, Sean
  Welleck, Yejin Choi, and Za{\"{\i}}d Harchaoui.
\newblock {MAUVE:} measuring the gap between neural text and human text using
  divergence frontiers.
\newblock In Marc'Aurelio Ranzato, Alina Beygelzimer, Yann~N. Dauphin, Percy
  Liang, and Jennifer~Wortman Vaughan (eds.), \emph{Advances in Neural
  Information Processing Systems 34: Annual Conference on Neural Information
  Processing Systems 2021, NeurIPS 2021, December 6-14, 2021, virtual}, 2021.

\bibitem[Radford et~al.(2019)Radford, Wu, Child, Luan, Amodei, Sutskever,
  et~al.]{radford2019gpt2}
Alec Radford, Jeffrey Wu, Rewon Child, David Luan, Dario Amodei, Ilya
  Sutskever, et~al.
\newblock Language models are unsupervised multitask learners.
\newblock \emph{OpenAI blog}, 1\penalty0 (8), 2019.

\bibitem[Ranzato et~al.(2016)Ranzato, Chopra, Auli, and
  Zaremba]{ranzato2015exposurebias1}
Marc'Aurelio Ranzato, Sumit Chopra, Michael Auli, and Wojciech Zaremba.
\newblock Sequence level training with recurrent neural networks.
\newblock In Yoshua Bengio and Yann LeCun (eds.), \emph{4th International
  Conference on Learning Representations, {ICLR} 2016, San Juan, Puerto Rico,
  May 2-4, 2016, Conference Track Proceedings}, 2016.

\bibitem[Robertson \& Zaragoza(2009)Robertson and Zaragoza]{Robertson2009ThePR}
S.~Robertson and H.~Zaragoza.
\newblock The probabilistic relevance framework: Bm25 and beyond.
\newblock \emph{Found. Trends Inf. Retr.}, 3, 2009.

\bibitem[Robertson et~al.(2009)Robertson, Zaragoza, et~al.]{robertson2009BM25}
Stephen Robertson, Hugo Zaragoza, et~al.
\newblock The probabilistic relevance framework: Bm25 and beyond.
\newblock \emph{Foundations and Trends{\textregistered} in Information
  Retrieval}, 3\penalty0 (4), 2009.

\bibitem[Salton \& Buckley(1988)Salton and Buckley]{salton1988idf}
Gerard Salton and Christopher Buckley.
\newblock Term-weighting approaches in automatic text retrieval.
\newblock \emph{Information processing \& management}, 24\penalty0 (5), 1988.

\bibitem[Sanh et~al.(2022)Sanh, Webson, Raffel, Bach, Sutawika, Alyafeai,
  Chaffin, Stiegler, Raja, Dey, Bari, Xu, Thakker, Sharma, Szczechla, Kim,
  Chhablani, Nayak, Datta, Chang, Jiang, Wang, Manica, Shen, Yong, Pandey,
  Bawden, Wang, Neeraj, Rozen, Sharma, Santilli, F{\'{e}}vry, Fries, Teehan,
  Scao, Biderman, Gao, Wolf, and Rush]{2021SanhT0}
Victor Sanh, Albert Webson, Colin Raffel, Stephen~H. Bach, Lintang Sutawika,
  Zaid Alyafeai, Antoine Chaffin, Arnaud Stiegler, Arun Raja, Manan Dey,
  M~Saiful Bari, Canwen Xu, Urmish Thakker, Shanya~Sharma Sharma, Eliza
  Szczechla, Taewoon Kim, Gunjan Chhablani, Nihal~V. Nayak, Debajyoti Datta,
  Jonathan Chang, Mike~Tian{-}Jian Jiang, Han Wang, Matteo Manica, Sheng Shen,
  Zheng~Xin Yong, Harshit Pandey, Rachel Bawden, Thomas Wang, Trishala Neeraj,
  Jos Rozen, Abheesht Sharma, Andrea Santilli, Thibault F{\'{e}}vry, Jason~Alan
  Fries, Ryan Teehan, Teven~Le Scao, Stella Biderman, Leo Gao, Thomas Wolf, and
  Alexander~M. Rush.
\newblock Multitask prompted training enables zero-shot task generalization.
\newblock In \emph{The Tenth International Conference on Learning
  Representations, {ICLR} 2022, Virtual Event, April 25-29, 2022}, 2022.

\bibitem[Seo et~al.(2019)Seo, Lee, Kwiatkowski, Parikh, Farhadi, and
  Hajishirzi]{seo2019real}
Minjoon Seo, Jinhyuk Lee, Tom Kwiatkowski, Ankur~P Parikh, Ali Farhadi, and
  Hannaneh Hajishirzi.
\newblock Real-time open-domain question answering with dense-sparse phrase
  index.
\newblock \emph{arXiv preprint arXiv:1906.05807}, 2019.

\bibitem[Shi et~al.(2024)Shi, Yang, Cai, Zhang, Wang, Yang, and
  Lam]{shi2024thorough}
Chufan Shi, Haoran Yang, Deng Cai, Zhisong Zhang, Yifan Wang, Yujiu Yang, and
  Wai Lam.
\newblock A thorough examination of decoding methods in the era of llms.
\newblock \emph{arXiv preprint arXiv:2402.06925}, 2024.

\bibitem[Shrivastava \& Li(2014)Shrivastava and Li]{shrivastava2014mips1}
Anshumali Shrivastava and Ping Li.
\newblock Asymmetric {LSH} {(ALSH)} for sublinear time maximum inner product
  search {(MIPS)}.
\newblock In Zoubin Ghahramani, Max Welling, Corinna Cortes, Neil~D. Lawrence,
  and Kilian~Q. Weinberger (eds.), \emph{Advances in Neural Information
  Processing Systems 27: Annual Conference on Neural Information Processing
  Systems 2014, December 8-13 2014, Montreal, Quebec, Canada}, 2014.

\bibitem[Su \& Collier(2022)Su and Collier]{su2022contrastive}
Yixuan Su and Nigel Collier.
\newblock Contrastive search is what you need for neural text generation.
\newblock \emph{arXiv preprint arXiv:2210.14140}, 2022.

\bibitem[Su et~al.(2022)Su, Lan, Wang, Yogatama, Kong, and
  Collier]{su2022metric2}
Yixuan Su, Tian Lan, Yan Wang, Dani Yogatama, Lingpeng Kong, and Nigel Collier.
\newblock A contrastive framework for neural text generation.
\newblock \emph{Advances in Neural Information Processing Systems}, 35, 2022.

\bibitem[Vaswani et~al.(2017)Vaswani, Shazeer, Parmar, Uszkoreit, Jones, Gomez,
  Kaiser, and Polosukhin]{vaswani2017Transformers}
Ashish Vaswani, Noam Shazeer, Niki Parmar, Jakob Uszkoreit, Llion Jones,
  Aidan~N. Gomez, Lukasz Kaiser, and Illia Polosukhin.
\newblock Attention is all you need.
\newblock In Isabelle Guyon, Ulrike von Luxburg, Samy Bengio, Hanna~M. Wallach,
  Rob Fergus, S.~V.~N. Vishwanathan, and Roman Garnett (eds.), \emph{Advances
  in Neural Information Processing Systems 30: Annual Conference on Neural
  Information Processing Systems 2017, December 4-9, 2017, Long Beach, CA,
  {USA}}, 2017.

\bibitem[Wang et~al.(2023)Wang, Song, Drozdov, Garimella, Manjunatha, and
  Iyyer]{wang2023knnlmdoesnot}
Shufan Wang, Yixiao Song, Andrew Drozdov, Aparna Garimella, Varun Manjunatha,
  and Mohit Iyyer.
\newblock Knn-lm does not improve open-ended text generation.
\newblock \emph{ArXiv preprint}, abs/2305.14625, 2023.

\bibitem[Welleck et~al.(2019)Welleck, Brantley, III, and
  Cho]{welleck2019selfreinforcement}
Sean Welleck, Kiant{\'{e}} Brantley, Hal~Daum{\'{e}} III, and Kyunghyun Cho.
\newblock Non-monotonic sequential text generation.
\newblock In Kamalika Chaudhuri and Ruslan Salakhutdinov (eds.),
  \emph{Proceedings of the 36th International Conference on Machine Learning,
  {ICML} 2019, 9-15 June 2019, Long Beach, California, {USA}}, volume~97 of
  \emph{Proceedings of Machine Learning Research}, 2019.

\bibitem[Welleck et~al.(2020)Welleck, Kulikov, Roller, Dinan, Cho, and
  Weston]{welleck2019metric1}
Sean Welleck, Ilia Kulikov, Stephen Roller, Emily Dinan, Kyunghyun Cho, and
  Jason Weston.
\newblock Neural text generation with unlikelihood training.
\newblock In \emph{8th International Conference on Learning Representations,
  {ICLR} 2020, Addis Ababa, Ethiopia, April 26-30, 2020}, 2020.

\bibitem[Weston et~al.(2018)Weston, Dinan, and Miller]{Weston2018RetrieveAR}
Jason Weston, Emily Dinan, and Alexander Miller.
\newblock Retrieve and refine: Improved sequence generation models for
  dialogue.
\newblock In \emph{Proceedings of the 2018 {EMNLP} Workshop {SCAI}: The 2nd
  International Workshop on Search-Oriented Conversational {AI}}, 2018.

\bibitem[Wu et~al.(2019)Wu, Wei, Huang, Wang, Li, and Zhou]{wu2019response}
Yu~Wu, Furu Wei, Shaohan Huang, Yunli Wang, Zhoujun Li, and Ming Zhou.
\newblock Response generation by context-aware prototype editing.
\newblock In \emph{The Thirty-Third {AAAI} Conference on Artificial
  Intelligence, {AAAI} 2019, The Thirty-First Innovative Applications of
  Artificial Intelligence Conference, {IAAI} 2019, The Ninth {AAAI} Symposium
  on Educational Advances in Artificial Intelligence, {EAAI} 2019, Honolulu,
  Hawaii, USA, January 27 - February 1, 2019}, 2019.

\bibitem[Ye et~al.(2023)Ye, Cao, Chen, Xu, and Zou]{ye2023fits}
Qichen Ye, Bowen Cao, Nuo Chen, Weiyuan Xu, and Yuexian Zou.
\newblock Fits: Fine-grained two-stage training for knowledge-aware question
  answering.
\newblock \emph{arXiv preprint arXiv:2302.11799}, 2023.

\bibitem[Yogatama et~al.(2021)Yogatama, de~Masson~d{'}Autume, and
  Kong]{yogatama2021adaptive}
Dani Yogatama, Cyprien de~Masson~d{'}Autume, and Lingpeng Kong.
\newblock Adaptive semiparametric language models.
\newblock \emph{Transactions of the Association for Computational Linguistics},
  9, 2021.

\bibitem[Zhang et~al.(2022)Zhang, Roller, Goyal, Artetxe, Chen, Chen, Dewan,
  Diab, Li, Lin, et~al.]{zhang2022opt}
Susan Zhang, Stephen Roller, Naman Goyal, Mikel Artetxe, Moya Chen, Shuohui
  Chen, Christopher Dewan, Mona Diab, Xian Li, Xi~Victoria Lin, et~al.
\newblock Opt: Open pre-trained transformer language models.
\newblock \emph{arXiv preprint arXiv:2205.01068}, 2022.

\bibitem[Zhang et~al.(2019)Zhang, Feng, Meng, You, and
  Liu]{zhang2019exposurebias2}
Wen Zhang, Yang Feng, Fandong Meng, Di~You, and Qun Liu.
\newblock Bridging the gap between training and inference for neural machine
  translation.
\newblock In \emph{Proceedings of the 57th Annual Meeting of the Association
  for Computational Linguistics}, 2019.

\bibitem[Zhong et~al.(2022)Zhong, Lei, and Chen]{zhong2022training}
Zexuan Zhong, Tao Lei, and Danqi Chen.
\newblock Training language models with memory augmentation.
\newblock In \emph{Proceedings of the 2022 Conference on Empirical Methods in
  Natural Language Processing}, 2022.

\end{thebibliography}
\bibliographystyle{iclr2024_conference}

\appendix
\section{Phrase Table Pruning and Phrase Matching}
\label{Appx: phrase table pruning and phrase matching}
It is noteworthy that syntactic parsing is a very well-studied task in NLP as well as its cross-domain and cross-language generalization. For example, the Universal Dependencies \footnote{https://universaldependencies.org/} project provides consistent grammatical annotation across over 100 languages. To our knowledge, the state-of-the-art parsing accuracies are pretty high for major languages such as English, Chinese, Italian, Japanese, Portuguese, etc. Nevertheless, we anticipate performance degradation for languages and domains when the parser accuracy is relatively low. For situations where a syntactic parser is unavailable, alternative methods may be utilized such as unsupervised syntactic parsing and unsupervised tokenization methods (\textit{e.g.}, BPE, sentencepiece).

After extracting constituents from the training data and supporting documents, we filter these constituents based on the following criteria: (1) remove trivial spans with the following constituent labels: `X', `PRT', `CC', `DT', `EX', `FRAG', `GW', `HYPH', `IN', `INTJ', `LS', `LST', `MD', `NFP', `NML', `PDT', `POS', `PP', `PRP', `PRP\$', `PPZ', `RB', `RBR', `RBS', `RP', `S', `SYM', `TO', `WDT', `WHADJP', `WHADVP', `WHNP', `WHPP', `WP', `WP\$', `WRB', `\#', `\$', `"', ```', `-LRB-', `-RRB-', `,', `.', `:'; (2) exclude constituents that are too short ($<2$ words) or too long ($>10$ words); (3) discard constituents with excessively high or low Inverse Document Frequency (IDF) values. The (minimum, maximum) thresholds for constituents with different numbers of words are: 2: (10.50, 14.08), 3: (11.09, 14.08), 4: (11.77, 14.30), 5: (12.10, 14.30), 6: (12.32, 14.30), 7: (12.51, 14.59), 8: (12.59, 14.59), 9: (12.64, 14.59), 10: (12.69, 14.59).
Next, we group lexically identical phrases and retrieve the top-$10$ candidates for each phrase using the BM25 algorithm \citep{robertson2009BM25}. We then calculate the semantic similarities between the original phrase and the retrieved candidate phrases using an off-the-shelf phrase encoder \citep{lee2021densephrase}. As a result, we can identify the most appropriate next phrase for each prefix based on the scores.

The entire preprocessing process, including syntactic parsing, phrase selection, and semantic matching, takes approximately 24 hours on 8 V100 GPUs. The overhead is small compared to the cost of training the model.

\section{Example for Iterative Self-Reinforcement}
\label{self-reinfo examples}
Suppose we have a prefix p = \textit{"Go right for the top when you"}. The ground truth for this prefix is \textit{"Go right for the top when you want to make things happen"}. The initial target phrase determined by the heuristics might be \textit{"want"}.
In the iterative self-reinforcement process, we would first let the model retrieve the \textit{k}-best phrases for the prefix from the entire candidate pool. Supposing that the \textit{k}-best phrases are [\textit{"want"}, \textit{"want to"}, \textit{"want to make things happen"}, \textit{"need"}, \textit{"can"}], only \textit{"want"}, \textit{"want to"}, and \textit{"want to make things happen"} are considered as valid ones. If the model's semantic matching score is highest for \textit{"want to make things happen"}, we would update the target phrase for the prefix to this phrase. If none of the \textit{k}-best phrases are valid, we will retain the previous target \textit{"want"}.
    
\section{Details of Task Phrasing and Specifications}
\label{sec:appendix}

% \begin{figure}[htb]
% \centering
% \includegraphics[width=0.47\textwidth]{./winogrande.png} 
% \caption{Formatted dataset example for Winograd-WSC. The format of Winogrande is the same as this.}
% \label{Fig.winogrande_example}
% \end{figure}

% \begin{figure}[htb]
% \centering
% \includegraphics[width=0.47\textwidth]{./openbookqa.png} 
% \caption{Formatted dataset example for OpenbookQA.}
% \label{Fig.openbookqa_example}
% \end{figure}

% \begin{figure}[htb]
% \centering
% \includegraphics[width=0.47\textwidth]{./truthfulqa.png} 
% \caption{Formatted dataset example for truthfulQA.}
% \label{Fig.truthfulqa_example}
% \end{figure}

% \begin{figure}[htb]
% \centering
% \includegraphics[width=0.47\textwidth]{./copa.png} 
% \caption{Formatted dataset example for COPA.}
% \label{Fig.copa_example}
% \end{figure}

% \begin{figure}[!ht]
% \centering
% \includegraphics[width=0.47\textwidth]{./wizard.png} 
% \caption{Formatted dataset example for Wizard of Wikipedia.}
% \label{Fig.wizard_example}
% \end{figure}
The statistics of the datasets we select are as follows:

\textbf{OpenbookQA} \citep{mihaylov2018openbookqa} is a collection of 5,957 multiple-choice questions, each with four options, centered around elementary scientific knowledge. We utilize the test split, which comprises 500 questions.

\textbf{ARC-Challenge} \citep{clark2018ARC} includes 7,787 authentic, grade-school level, multiple-choice science questions. These questions span a wide range of topics in science and history, among others. Our experiments focus on the test split of its Challenge Set, which contains 1,172 hard questions.

\textbf{TruthfulQA} \citep{lin2021truthfulqa} is a distinctive dataset emphasizing the truthfulness of answers. We employ the test split of the multiple-choice option, which includes 817 questions.

\textbf{MedMCQA} \citep{medmcqa} is a comprehensive, high-quality dataset designed for biomedical question-answering. We use its validation split, which consists of 4,183 questions.

\textbf{Med-USMILE} \citep{jin202medusmile} encompasses 12,723 multiple-choice questions, each with four options, originally sourced from the National Medical Board Examination in the USA. We utilize its test split, which includes 1,273 questions.

Given a question with several candidate answers, we concatenate the question with each candidate answer to form options, and then ask the model to select the most accurate one among all the options.
We remove questions from these datasets where all candidate answers are single words to ensure the inclusion of phrases in the retrieval process.

% Figure \ref{Fig.openbookqa_example}, \ref{Fig.truthfulqa_example} illustrate the formatting and phrasing of all the tasks included in the paper. 
\section{Case Study}
\label{Appx: case study}
\begin{figure*}[t]
\centering
\includegraphics[width=\textwidth]{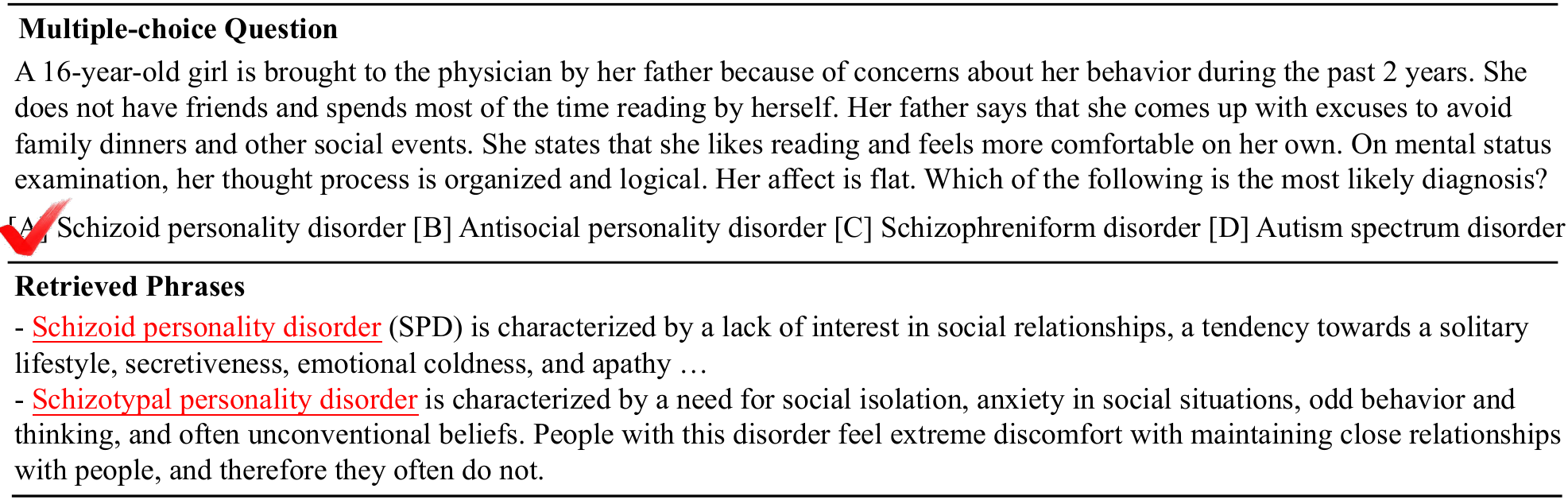} 
\caption{An illustrative example from Med-USMILE: The two highlighted phrases in red are retrieved in response to the posed question.}
% , underscoring our model's adeptness at harnessing information-rich phrases.}
\label{Fig.QA_example}
\end{figure*}
To elucidate the role of phrase retrieval in knowledge-intensive tasks, we delve into a case study depicted in Figure \ref{Fig.QA_example}. 
As previously discussed in Section \ref{Text Probabilities Calculation}, our approach involves retrieving phrases for each token in an option, enabling us to estimate the probabilities of alternative generation paths beyond simply generating the token sequence.
In this specific case from the Med-USMILE dataset, options are formed by concatenating the question with each candidate answer. We find that the phrases retrieved for the final token of the question include the answer, a proper noun requiring medical knowledge for understanding. This introduces a new generation path: $\text{question}\rightarrow \textit{Schizoid personality disorder}$.
% , \textit{Schizoid personality disorder}, which is a proper noun that necessitates medical knowledge for comprehension, introducing a new generation path: $\textbf{question}\rightarrow $
We observe that the contexts of the retrieved phrases, such as ``\textit{Schizoid personality disorder (SPD) is characterized by a lack of interest in social relationships ...}'', align closely with the context of the question, ``\textit{She does not have friends and spends most of the time reading by herself ...}''. 
These contextually encoded phrases benefit answer selection, thereby showcasing the interpretability of our model.
It also highlights the model's ability to leverage contextual information effectively, particularly in tasks that require specialized knowledge.
\section{Details for automatic evaluation metrics}
\label{Appx: metrics}
In this section, we provide a detailed introduction to MAUVE, as well as the concepts of coherence and diversity.

\textbf{MAUVE} \citep{pillutla2021mauve} measures how closely the token distribution in generated text matches that in human-written text across the entire test set.

\textbf{Coherence} \citep{su2022contrastive, su2022metric2} measures the semantic coherence between the prompt $x$ and the generated text $\hat{x}$ by calculating the average log-likelihood as: $\text{coherence}(\hat{x}; x) = \frac{1}{|\hat{x}|} \sum_{i=1}^{|\hat{x}|} \log p_M(\hat{x}_i | [x : \hat{x}_{<i}])$, where $[:]$ is the concatenation operation and $M$ is a pre-trained LM. We follow prior work and set $M$ as the OPT-2.7B model \citep{zhang2022opt}. In our implementation, we introduce a slight modification by taking the negative of the average log-likelihood. This adjustment transforms the typically negative log-likelihood into a positive value, facilitating a more intuitive interpretation of the results.

\textbf{Diversity} \citep{welleck2019metric1,su2022metric2,lan2023cog} measures the repetition in generated text at different $n$-gram levels by computing the proportion of unique $n$-grams to total $n$-grams in the generated text. It is defined as: $\text{diversity}=\prod_{n=2}^4(1.0-\frac{\text{rep-n}}{100})$, where $\text{rep-n}=100\times(1.0-\frac{|\text{unique n-grams}(\hat{x})|}{|\text{total n-grams}(\hat{x})|})$, and $\hat{x}$ is the text generated by the model.

\begin{table*}[t]
\normalsize
\centering
\setlength{\tabcolsep}{4pt}
\begin{tabular}{ccccccc}
\toprule
$k$ & \textbf{TruthfulQA} & \textbf{OpenbookQA} & \textbf{ARC-Challenge} & \textbf{MedMCQA} & \textbf{Med-USMILE} & \textbf{Avg.} \\
\hline
1 & 32.74 & \textbf{36.80} & 27.94 & \textbf{29.95} & 25.68 & 30.62 \\
2 & 32.88 & 36.80 & 28.04 & 29.90 & 25.68 & 30.66 \\
4 & 33.29 & 36.80 & 27.84 & 29.84 & 25.68 & 30.69 \\
8 & 33.42 & 36.80 & 27.84 & 29.76 & 25.42 & 30.65 \\
16 & 34.25 & 36.80 & 27.64 & 29.61 & 25.15 & 30.69 \\
32 & 34.11 & 36.27 & 27.64 & 29.50 & 26.12 & 30.73 \\
48 & \textbf{34.38} & 36.27 & 28.04 & 29.27 & \textbf{26.21} & \textbf{30.83} \\
64 & 33.84 & 36.53 & \textbf{28.34} & 29.38 & 25.59 & 30.74 \\
128 & 34.27 & 36.27 & 28.24 & 29.44 & 25.69 & 30.78 \\
256 & 33.42 & 36.27 & 27.37 & 29.24 & 24.80 & 30.22 \\
512 & 32.88 & 35.73 & 27.64 & 29.33 & 25.68 & 30.25 \\
768 & 32.47 & 35.47 & 27.74 & 29.67 & 25.42 & 30.15 \\
1024 & 32.47 & 35.47 & 27.54 & 29.61 & 24.89 & 30.00 \\
\bottomrule
\end{tabular}
\caption{\label{ablation-k}
Ablation studies on the impact of $k$ on knowledge-intensive tasks.}
\end{table*}

\begin{table*}[t]
\normalsize
\centering
\setlength{\tabcolsep}{2.5pt}
\begin{tabular}{cccccc}
\toprule
& \textbf{TruthfulQA} & \textbf{OpenbookQA} & \textbf{ARC-Challenge} & \textbf{MedMCQA} & \textbf{Med-USMILE} \\
\hline
Base LM & 30.14 & 22.40 & 22.41 & 28.27 & 23.58 \\
$k$NN-LM & 30.14 & 22.40 & 23.32 & 27.99 & 23.14 \\
COG & 32.88 & 34.13 & 25.13 & 29.16 & 25.15 \\
Ours & \textbf{33.29} & \textbf{35.20} & \textbf{27.04} & \textbf{30.24} & \textbf{26.21} \\
Ours (w/o phrase) & 28.22 & 21.87 & 23.02 & 27.99 & 24.89 \\
\bottomrule
\end{tabular}
\caption{\label{from scratch result}
The results of models trained from scratch.}
\end{table*}

\begin{table*}[t!]
\normalsize
\centering
\setlength{\tabcolsep}{6pt}
\begin{tabular}{cccccc}
\toprule
& \textbf{TruthfulQA} & \textbf{OpenbookQA} & \textbf{ARC-Challenge} & \textbf{MedMCQA} & \textbf{Med-USMILE} \\
\hline
w/o SR & 34.11 & \textbf{37.07} & 27.14 & \textbf{30.32} & \textbf{25.85} \\
round1 & 33.97 & 36.80 & 27.34 & 29.84 & 25.77 \\
round2 & \textbf{34.27} & 36.27 & \textbf{28.24} & 29.44 & 25.69 \\
\bottomrule
\end{tabular}
\caption{\label{ablation-SR-QA}
Ablation studies on the effect of self-reinforcement.}
\end{table*}

\section{Detailed Human Evaluation}
\label{Appx: Detailed Human Evaluation}
Following the results in Table \ref{table:human eval}, upon manual analysis, we find that the outputs from our method often have a tighter connection with the preceding text (coherence) and exhibit stronger knowledge characteristics (informativeness). For instance, they often include specialized terms, which are not observed in the outputs from the base LM. As for the lower scores of the base LM compared to the base LM (w/o FT), one major reason is that the fine-tuned model frequently outputs "References: xxx" and "External Links: xxx". This is related to the characteristics of the Wikipedia dataset, where each article typically ends with references and external links. To further investigate this, we retest the Base LM, excluding all outputs containing "References" and "External Links" (which accounts for 26.77\% of the cases). The resulting MAUVE score is 60.23, slightly lower than the 69.68 of the base LM (w/o FT), but substantially higher than the previous score of 42.61. This also suggests that one possible reason for the significant drop in the MAUVE score after (\textit{c.f.} Table \ref{open-ended generation}) fine-tuning the base model is due to these extraneous outputs.

\section{Ablation studies}
\label{Appx: ablation studies}

\paragraph{The impact of $k$.}
As shown in Table \ref{ablation-k}, $k$ does not have a significant impact on the performance of our model on knowledge-intensive tasks. Since we retrieve the top $128$ phrases during the self-reinforcement process, we set $k=128$ throughout all experiments.

\paragraph{The impact of pre-trained models.}
The results are given in Table \ref{from scratch result}. Our model outperforms the base LM across all datasets, achieving a 12.8\% absolute improvement on OpenbookQA.
This suggests that our training framework is not heavily dependent on pre-trained models.

\begin{wrapfigure}{r}{0.45\textwidth}
\vspace{-3.5mm}
\centering 
\includegraphics[width=6.3cm]{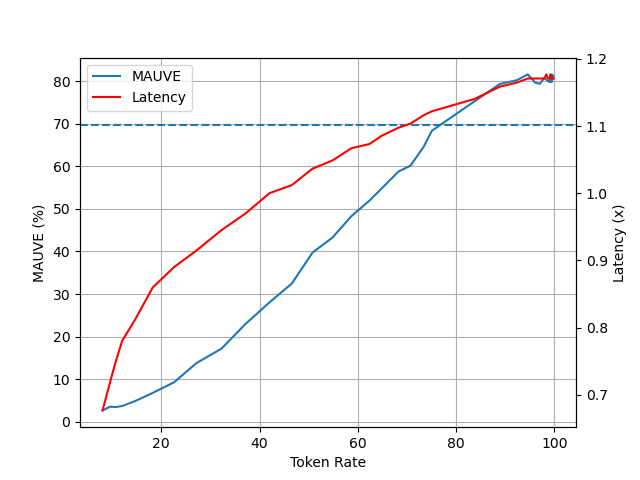}
\caption{The MAUVE score and latency of our model with different token rates.}
\label{Fig.mauve}
\end{wrapfigure}

\paragraph{The impact of phrase retrieval threshold $\phi$.}
\label{analysis: threshold}
The phrase retrieval threshold, which filters out phrases with probabilities below it, influences the proportion of tokens (token rate for short) used in the text generation process. This section explores the intriguing relationship between token rate and the quality of the generated text.

A lower token rate boosts the model's inference efficiency, as phrases typically contain multiple tokens, thereby reducing the number of decoding steps. However, this efficiency can sometimes compromise the quality of the text. This degradation occurs due to the inherent uncertainty during generation, particularly when all candidate probabilities are low or when an inappropriate candidate is sampled. In such cases, the selection of an incorrect phrase, given its length, can significantly disrupt the generation process, a phenomenon known as exposure bias \citep{ranzato2015exposurebias1,zhang2019exposurebias2}. Conversely, the impact of choosing a sub-optimal token is less severe.
Therefore, up to a certain point (approximately 0.89), increasing the token rate can stabilize the generation process, as shown in Figure \ref{Fig.mauve}. Beyond this point, the model's performance peaks. 

Furthermore, even without phrases (\textit{i.e.}, the token rate is 1), our model can generate high-quality text, suggesting that our method also enhances token prediction learning.
However, it's important to note that while our model achieves a high MAUVE score based solely on token prediction, the factuality of the generated text is lower than when phrase retrieval is integrated. This highlights the need for more innovative metrics to precisely measure the quality of generated text.

\paragraph{The impact of self-reinforcement on knowledge-intensive tasks.}
Based on the results presented in Table \ref{ablation-SR-QA}, it can be concluded that the utilization of SR does not significantly affect the performance of our model on knowledge-intensive tasks. This implies that our framework is inherently capable of effectively addressing such tasks, even in the absence of SR.

\section{Limitations and Future Work}
\label{limitation}
While our proposed framework has shown promising results in efficiently generating accurate and coherent text, it is important to acknowledge the limitations of its current form. First of all, the presented result is just a proof-of-concept of the new paradigm. Future work should focus on (1) Scalability. In our current experiments, we train our models and build the phrase index on the English Wikipedia corpus. When scaling up to larger corpora, we may encounter computational challenges due to the significantly increased amount of possible phrases. To make our approach scalable, some potential solutions include clustering, dimensionality reduction, and fast vector search algorithms with sub-linear time complexity. (2) Alignment. Recent studies have shown that alignment (i.e., fine-tuning language models to following human instructions) is important to make language models universally useful. Thus, incorporating alignment techniques into our approach is an important future research direction.
% (2) The quality and coverage of the phrase index. If the phrases in the index are not diverse enough or do not cover certain fields, the model may struggle to generate accurate and coherent outputs for those topics. We are actively working on addressing these issues and improving our model's performance in these challenging scenarios.
%Orthogonal to our exploration of a new paradigm for retrieval augmentation, we are also excited to observe the emergence of new methods in the development of retrieval systems. For example, \citet{jin2023diffusionret} introduce a diffusion-based retrieval framework, which gradually generates a joint distribution from noise to model the retrieval task. \citet{li2023freestyleret} focus on a style-diversified query-based retrieval task to satisfy various user intentions. These works provide fresh perspectives and inspire further advancements in retrieval methods.
\end{document}